\documentclass[letter, 10pt, conference]{ieeeconf}      

\IEEEoverridecommandlockouts                              
\usepackage{amsmath}
\usepackage{color}
\definecolor{chred}{rgb}{0.8,0,0}
\usepackage{graphicx}
\usepackage{subfigure}
\usepackage{eqlist}
\usepackage{txfonts}
\usepackage{footmisc}
\usepackage{booktabs}
\usepackage[lined, ruled, linesnumbered, commentsnumbered]{algorithm2e}

\title{\LARGE \bf Developing and Comparing Single-arm and Dual-arm Regrasp}

\author{Weiwei Wan and Kensuke Harada
  \thanks{Weiwei Wan and Kensuke Harada are affiliated with the 
  Manipulation Research Group,
  National Institute of Advanced Industrial Science and Technology (AIST).
  {\tt\small wan-weiwei@aist.go.jp}}
}

\begin{document}

\maketitle
\thispagestyle{empty}
\pagestyle{empty}

\begin{abstract}

The goal of this paper is to develop efficient regrasp algorithms for single-arm
and dual-arm regrasp and compares the performance of single-arm and
dual-arm regrasp by running the two algorithms thousands of times. We focus on
pick-and-place regrasp which reorients an object from one placement to another
by using a sequence of pick-ups and place-downs. After analyzing the simulation
results, we find dual-arm regrasp is not necessarily better than single-arm
regrasp: Dual-arm regrasp is flexible.
When the two hands can grasp the object with good clearance, dual-arm regrasp
is better and has higher successful rate than single-arm regrasp.
However, dual-arm regrasp suffers from geometric constraints caused by the two
arms. When the grasps overlap, dual-arm regrasp is bad. Developers need to
sample grasps with high density to reduce overlapping. This leads to exploded
combinatorics in previous methods, but is possible with the
algorithms presented in this paper.
Following the results, practitioners may
choose single-arm or dual-arm robots by considering the object shapes and
grasps. Meanwhile, they can reduce overlapping and implement practical dual-arm
regrasp by using the presented algorithms.

\end{abstract}

\section{Introduction}

Regrasp can be performed by a single robotic arm plus an
extrinsic table surface\cite{Wan2015a}. The single arm changes the grasps
taking advantage of the table. It picks up the object
from its initial placement on the table, reorientates it
to a new state and place it down on the table, and
changes the grasps to pick it up again from the new state.
In this way, a single arm can finish difficult reorientation
tasks like flipping where the object cannot be directly
placed down into the goal state. 

Regrasp can also be performed by a dual-arm robot\cite{Jean10}.
The robot changes the grasps by handling the object
from one hand to another. One arm of the robot picks up
the object from its initial placement on the table,
reorientates it to a handling configuration in the work
space and hold it. The other arm grasps the object from the
handling configuration and reorientates it again.
In this way, dual-arm robots can also finish difficult
orientation tasks.

We wonder which has higher performance in pick-and-place regrasp.
We develop efficient regrasp algorithms for both
single-arm and dual-arm regrasp and compare their successful rates and time cost
by running the algorithms thousands of times. We demonstrate the algorithms
developed in this paper are fast enough for thousands of simulation and
concluded that dual-arm regrasp is not necessarily better than single-arm regrasp.
by analyzing the simulation results.
The performance of dual-arm regrasp depends on object shapes and the overlapping
of the grasps from the two hands. Practitioners may choose
single-arm or dual-arm robots by considering the object shapes and grasps.
Meanwhile, they can reduce overlapping and implement practical dual-arm regrasp by using
the algorithms presented here.

While this paper concentrates on the algorithm development and the comparison of
the single-arm and dual-arm regrasp, the results are also expected to be used
to automatically decide arm numbers: Robots can use the algorithms to
to predict whether she should use a single arm or dual arms.
%

\section{Related Work}


\subsection{Regrasp}

Most of the early work on regrasp planning uses a single arm.
The seminal study is done by Tournassound et al.\cite{Pierre87},
and is later described in detail in the Handey system\cite{Handeybook}.
This early study builds a Grasp-Placement (GP) table to search
for regrasp sequences. It solves the IK and checks all collisions
to invalidate the grasp and placement pairs, fill up the GP table,
and search the table to find a sequence of pick-and-place motion.
There are lots of work following this study. For example, Rohrdanz
et al.\cite{Rohrdanz97} improves the efficiency of regrasp planning using
an evaluated breadth-first search and rated grasp and placement
qualities. Terasaki et al.\cite{Hajime98} adds a simple rotating
mechanism to the robotic gripper, making the regrasp planning dynamic.
Their study not only regrasps objects using pick-and-place, but also
pivoting \cite{Goldberg94}. Stoeter et al.\cite{Sascha99} replaces the GP
table with a space of compatible grasp-placement-grasp triplets and search this
space to find a sequence of pick-and-place motion. More practically,
Cho et al.\cite{Cho03} implements the regrasp algorithm with a real
robot using off-line mapping and on-line retrieving. During
the off-line mapping, they sample the workspace and pre-build
a look-up table to hold the IK-feasible grasps at the
sampled positions. During the on-line retrieving, they check
the look-up table to quickly know whether regrasp is
feasible. Using the look-up table avoids explicit IK
solving and improves efficiency.

More recent work studies regrasp in the context of
hierarchical TAsk and Motion Planning (TAMP) and solves the
constraint satisfaction problem. The framework is presented
by Lozano-Perez et al.\cite{Tomas14} where a symbolic planner plans
a sequence of high-level operations and a Constraint Satisfaction
Problem (CSP) solver solves low-level problems. Under this
framework, regrasp planning is divided into the symbolic
pick-and-place sequence and the geometrically feasible placements,
paths, grasps, and locations. A robot decides high-level
operations using the symbolic planner and decides low-level
operations using the CSP solver. Lozano-Perez demonstrates
the framework using a single-arm of a PR2 robot.
The most challenging part of this framework is the exploded
low-level combinatorics (see Dogar et al.\cite{Dogar15}).
Properly rating regrasps is essential to use this framework to solve problems
with large constraint graphs.
A similar framework is presented by Lagriffoul et al.\cite{Lagriffoul12}.
Both single-arm and dual-arm regrasp
are performed to demonstrate the framework.

In our previous work, we develop a single-arm regrasp
algorithm using a novel graph and use the algorithm
to analyze the utility of tilted work surfaces\cite{Wan2015a}.
This paper continues the development by extending the algorithm
to dual-arm robots. It uses the regrasp algorithms to compare
the performance of single-arm and dual-arm regrasp.

\subsection{Dual-arm regrasp}

The difficulty of dual-arm regrasp is the high-dimensional
configuration space composed by the two arms and the exploded
number of combinatorics between the two grasps during handling.
The seminal work discussing this difficulty is by
Koga et al.\cite{Yoshihito92}. The paper compares three 2D demos --
two 2-DoF arms without obstacles, two 2-DoF arms with obstacles,
and two 3-DoF arms with obstacles. The first demo can be
exhaustively computed. The second demo employs manipulation
graph et al.\cite{Alami90} to make computation feasible. The third one
uses random sampling to further improve efficiency.

Following this initial study, lots of research are devoted
to the dual-arm regrasp problem. Koga himself extends the
2D work to 3D regrasp planning using two and three manipulators
and a small amount of manually specified grasp assignments\cite{Yoshihito94a}.
The extension shows multi-robot regrasp is possible using
the limited computational resources at that time.
Saut et al.\cite{Jean10} studies the dual-arm regrasp problem by using
an optimized regrasp position and object orientation.
The optimized regrasp position minimizes wrist motions
and the optimized object orientation
minimizes the approaching angles of the two
hands. The work is based on a roadmap composition work
introduced by Gharbi et al.\cite{Gharbi09}. It fails to output a
solution in some situations. Balaguer et al.\cite{Balaguer12}
studies the dual-arm regrasp problem by estimating the
two grasping positions for the two hands in the object?s
local coordinate system. It optimizes the regrasp position
and object orientation by minimizing the time needed to move
the two hands to the estimated positions. Approaching directions of the
hands are optimized later. Comparing with
Saut, the algorithm runs at only one object orientation,
but with more hand approaching directions. Vahrenkamp et al.\cite{Niko09}
studies the dual-arm regrasp problem by pre-building both a
single-arm manipulability map and a bi-manual manipulability
map in work space. They query the single-arm data map to
find the IK-feasible grasps of each hand and query the bi-manual
data map to assign scores to the possible two-hand combinations.
Their work is resolution-complete and can perform on-line search
in 20$ms$. Most up-to-date, Suarez et al.\cite{Suarez15} proposes employing
the synergy analysis which was initially discussed and used in
robotic grasping by Santello et al.\cite{Santello98} to reduce the
computational cost of dual-arm regrasp. They compute a PMDs
(Principle Motion Directions) manifold in the configuration
space of the dual-arms, and choose the grasp configurations
that have smaller distance to the PMDs manifold to reduce dual-arm
combinatorics. Interestingly, this study has the advantage in getting a
human-friendly robot motion.

Our dual-arm regrasp algorithm is based on the regrasp graph.
It is somewhere between \cite{Jean10} and \cite{Niko09},
and can be used on-line. In detail, we use manipulability
and approachability to find an optimized position and use sampling to
generate a sequence of orientation.
Each position and orientation pair is a handling configuration,
and the dual-arm robot performs regrasp at it. Our
algorithm is complete and fast enough to find a motion
sequence for difficult regrasp tasks like flipping.
Most importantly, we compare single-arm and dual-arm regrasp
by running the algorithms thousands
of times on random initial and goal states.
To the best of our knowledge, this is the first work that
compares the performance of single-arm and dual-arm regrasp.

\section{Overview of the Algorithms}

In this section, we present the algorithm flowcharts of single-arm and dual-arm
regrasp and discuss briefly the role of each part in the
flowcharts. The details will be discussed in later sections.

\begin{figure}[!htbp]
  \centering
  \includegraphics[width = 3.2in]{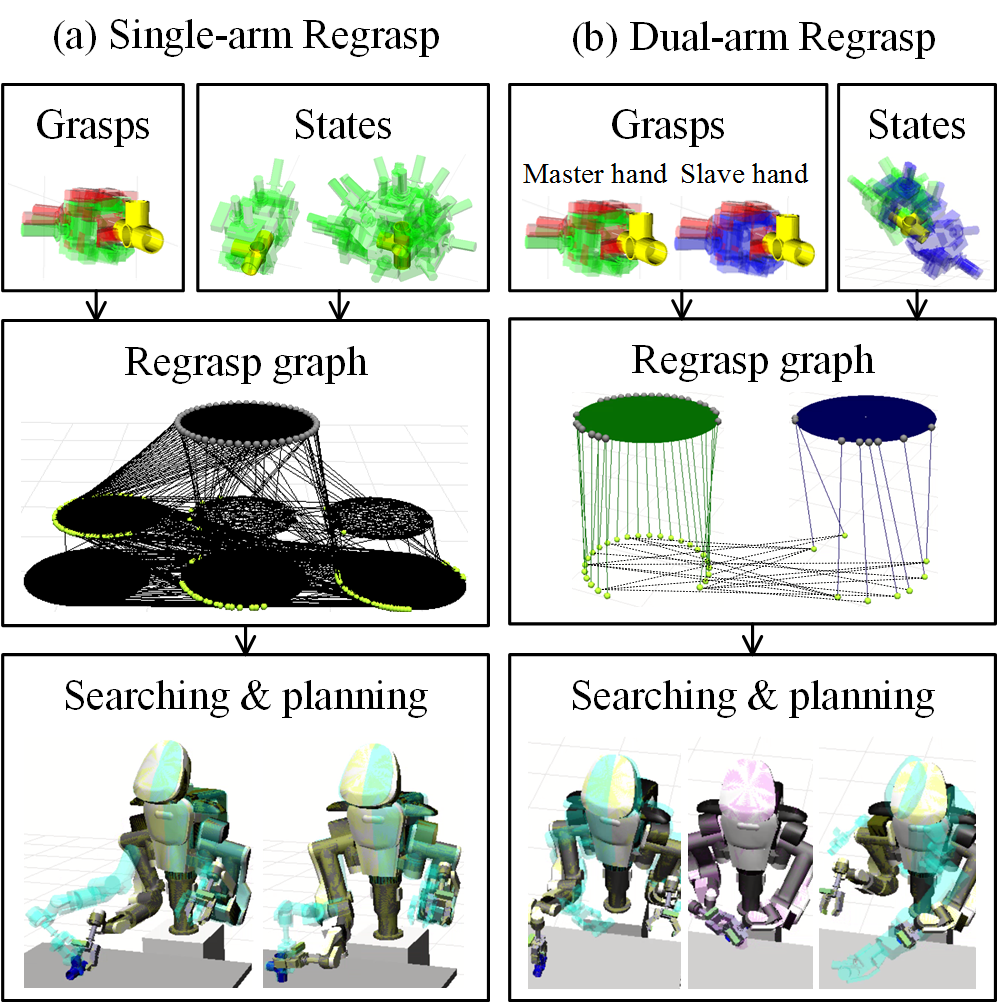}
  \caption{Flow charts of the two regrasp algorithms developed for
  single-arm and dual-arm regrasp. In the single-arm case,
  we first plan the grasps and states (placements) using the object model
  (rendered in yellow)
  and the robot's hand model (rendered in red or green,
  depending whether the hand and the object collide with each other or not).
  Then, we build a regrasp graph using the states and their associated
  grasps. Third, we 
  search the regrasp graph to find a regrasp
  sequence and do motion planning.
  The dual-arm case follows the same flow but
  involves two hands and a merged regrasp graph.
  The green hand and sub-graph denote the master and
  the blue hand and sub-graph denote the slave.
  The master hand handles the object to
  the slave hand.}
  \label{algorithmflow}
\end{figure}

\subsection{Single-arm Regrasp}

The flowchart of single-arm regrasp is shown
in Fig.\ref{algorithmflow}(a). In the first step, we compute the grasps
and states used for regrasp planning. The input of this step
is the model of a robotic gripper and an object. The output
includes (1) the resolution complete force-closure
grasps of the robotic gripper on the object, and (2) the
stable placements of the object on a table surface and the
grasps associated with each placement. The component that
computes the grasps is shown in the box named ``Grasps''.
The component that computes the placements and its
associated grasps is shown in the box named ``States''.
In the second step, we build a regrasp graph using the
grasps and states computed in step one. Each circle in
the graph represents one placement of the object, and
each node on the circle represents one grasp. The
edges encode the relationship between the grasps and
the placements: One can be changed into another by the robot
using transfer or transit motions\cite{Harada14b}.
The last step is to search the regrasp graph and
do motion planning. The graph search finds
is a sequence of pick-and-place operations and is in
the high level of task planning. The motion planning
finds a sequence of arm trajectories that
connects adjacent pick-and-place
operations in the high-level sequence.

\subsection{Dual-arm Regrasp}

The flowchart of dual-arm regrasp is shown
in Fig.\ref{algorithmflow}(b). Like the single-arm regrasp,
the dual-arm regrasp also includes a grasp-and-state planning step, a
regrasp graph building step, and a searching and motion planning
step. The difference is:
First, both hands are considered in the first step. The
grasp planning computes the collision-free grasps of both hands.
The state planning component associates the available grasps
of both hands to each placement.
Second, the state planning not only compute the placements
on a table, but also the handling configurations in the air. The
handling configurations are the positions and orientations of the
object for dual-arm handling. The state planning uses
the approachability and manipulability to find an optimized
position and uses sampling to generate a sequence
of rotation. Each position and rotation pair is
a candidate handling configuration. The state planning
associates the grasps of both hands to the handling configurations. 
Collision detections between hands are performed during the association.
Third, the regrasp graph is composed of two sub-graphs,
one for the master hand and one for the slave hand.
The circles in the higher-layer of each
sub-graph are the initial and goal placements
of the object on the table. The circles in
the lower-layer of each sub-graph are the handling configurations.
The two sub-graphs are connected at the lower-layer.
It is where the master hand
handles the object to the slave hand.

\section{Grasps and States}

\subsection{Grasp planning}

\begin{figure}[!htbp]
  \centering
  \includegraphics[width = 3.2in]{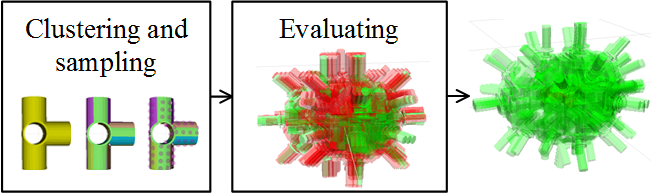}
  \caption{Grasp planning in the local coordiante system
  of the object. In the left, the object surface is clustered and sampled.
  In the middle, each pair of the sampled points
  from two parallel clusters is evaluated using grasp stability,
  force-closure, and collision. In the right, low-quality grasps are removed
  (denoted in red in the middle figure).}
  \label{graspplan}
\end{figure}

In detail, our grasp planning is done in the local coordinate system of the
object. The flow is shown in Fig.\ref{graspplan}.

First, we cluster the mesh model of the object and
sample contact points. The clustering will merge coplanar triangles
on the mesh model using a tolerance value given by the user.
On each cluster, we sample contact points using its
two principal axes. Suppose the bounding box of the cluster
is \{(${e_1}^{min}$, ${e_1}^{max}$), (${e_2}^{min}$, ${e_2}^{max}$)\}
and the two principal axes are \{$x_1$, $x_2$\},
we sample the surface following
$p=\omega_1x_1+\omega_2x_2$, $\omega_1\in[{e_1}^{min}$, ${e_1}^{max}]$,
$\omega_2\in[{e_2}^{min}$, ${e_2}^{max}]$.
The step is shown in the box named ``Clustering and sampling'' in
Fig.\ref{graspplan}.
Then, we evaluate each pair of contact pairs
from two parallel clusters by checking the grasp
stability, force-closure, and collision. For grasp
stability, we remove the
contact points that has a small offset to cluster boundary and low
resistance to external torques. For force-closure,
we measure each two contact points on parallel
clusters with eight approaching directions,
check if it has large wrench space ball, and remove the grasps
that have low quality. For collision,
we remove the grasps that induce collision between
fingers and collision with the object. This step is shown in the box
named ``Evaluation'' in Fig.\ref{graspplan}. The removed
grasps are drawn in red in the figure. In contrast,
the remaining grasps are drawn in green.

The output of grasp planning is the grasps in the object's local
coordinate system. Only the grasps of one hand is computed in
single-arm regrasp. The grasps of both hands are computed
in dual-arm regrasp.

\subsection{State planning}

\begin{figure}[!htbp]
  \centering
  \includegraphics[width = 3.2in]{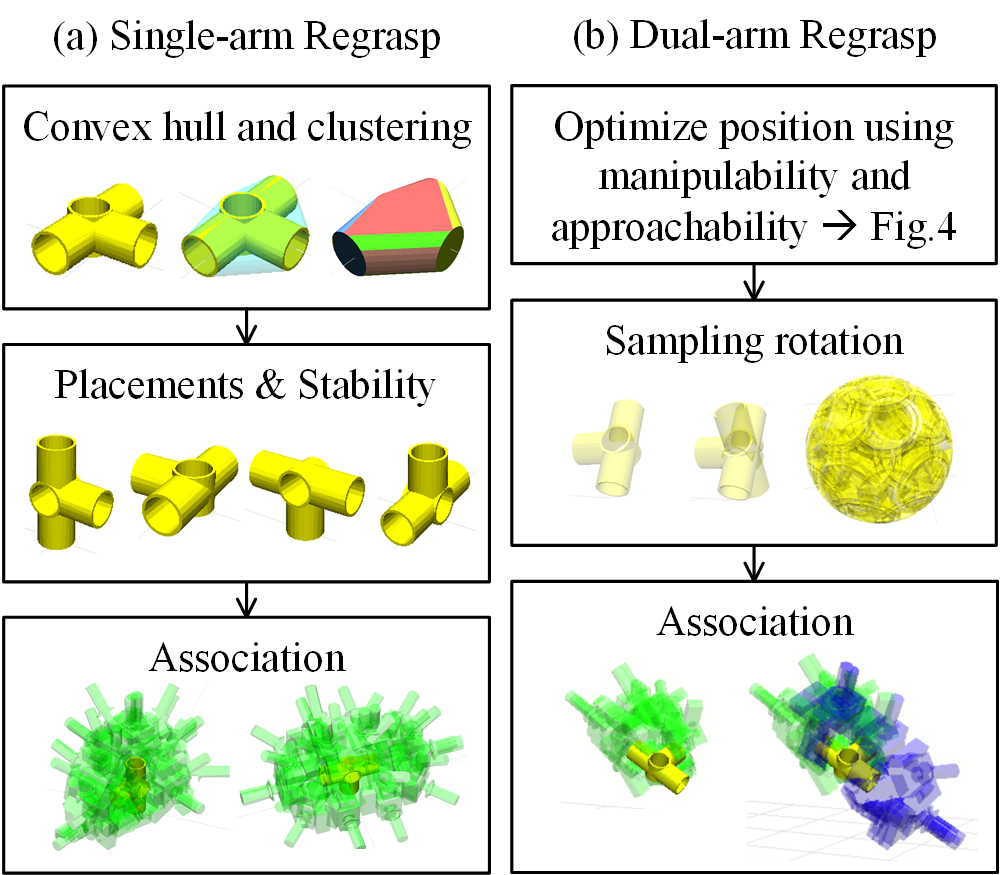}
  \caption{Computing the stable placements for
  single-arm regrasp and both the stable placements
  and the handling configurations for dual-arm regrasp.
  In the single-arm case, we cluster the convex hull of the object
  (top box) and test if any of the clusters could be the
  standing surface of a stable
  placement (middle box). Each stable placement is associated with
  available grasps (rendered in green in bottom box). In the 
  dual-arm case, we fix the object to an optimized position
  and sample the orientation (middle box).
  Each position and rotation pair is a candidate handling
  configuration and is associated with the 
  available grasps of both hands
  (rendered in red and
  blue in bottom box).}
  \label{stateplan}
\end{figure}

There are two kinds of states. One is the stable
placements of the object on a table surface, used in both
single-arm and dual-arm regrasp. The other one is the handling
configurations of the object in the air, used
only in dual-arm regrasp.
The flow charts of computing these two kinds of states are shown
in Fig.\ref{stateplan}(a) and (b) respectively.

\subsubsection{Placement planning}

The placement planning includes three steps. In the first step,
we compute the convex hull of the original object mesh and
perform surface clustering on the convex hull. The
surface clustering algorithm is the same as the one used in
grasp planning. Each cluster is one candidate standing
surface and the object may be placed on it.
Then, we test the stability of the objects standing on
these candidate surfaces. This is a filtering
process and the unstable placements, e.g. the placements
where the Zero Moment Point (ZMP) is outside the candidate
surface or too near to its boundary, will be removed.
In the third step, we
associate the grasps to the stable placements filtered by step two:
(1) We remove the grasps that collide with the table surface
and (2) we transform from the object?s local coordinate
system to each placement?s local coordinate system.

The placements and their associated grasps are described
in local coordinate systems. We perform collision detection
to ensure the hand, the object, and the table surface do not
collide with each other. However, the yaw of the object on
the table surface and the IK of the robot arm are not computed.
They are delayed to regrasp graph searching.

\subsubsection{Handling configuration planning}

\begin{figure}[!htbp]
  \centering
  \includegraphics[width = 3.2in]{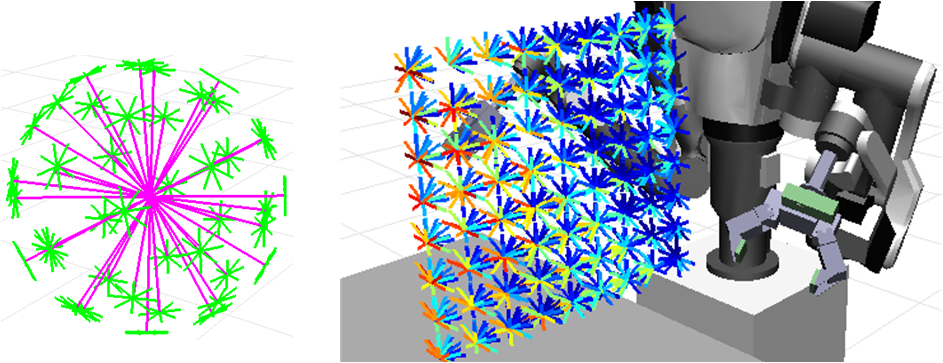}
  \caption{Optimizing the handling position using manipulability
  and approachability. First, we discretized the middle surface in front of the
  robot with lattice (see right figure). Then, for each grid, we sample the
  approaching directions pointing at it (see the purple vectors in left
  figure). Third, we sample the rotation around each approaching direction
  (see the green vectors at the end of each purple vector). A list of
  configurations is generated through the discretization.
  We compute the manipulability at each configuration, count the number of
  rotation with good manipulability at each grid,
  and save the value as the approachability. The grid with
  highest approachability is used as the optimal handling position.}
  \label{discretization}
\end{figure}

We use an heuristic method to plan the handling configurations
for handling. First, an optimized position is computed using
manipulability and approachability. This step involves
three levels of discretization. In level 1, we sample the
surface in the middle of the two arms with grids. In level 2,
at each of the sampled grid, we sample the approaching directions
pointing at it. In level 3, we sample the rotation around each
sampled approaching direction. After the three levels of
discretization, we get a list of rotation matrices representing
the configurations of the hand to grasp something on the middle
surface. We move the hand to these configurations and
compute the manipulability of the arm
using $\sqrt{\det{(\mathbf J \mathbf J^T)}}$
\cite{Yoshikawa85}.
There are two possible cases. One is IK-infeasible configuration.
The robot cannot approach the grid using the rotation matrix in this case and
the manipulability is 0. The other is the IK feasible configuration where the
manipulability has a positive value. If the manipulability of one
rotation matrix is not 0 and is larger than a threshold value,
we count it as approachable. At each sampled grid, we count the
number of approachable rotation matrices and use the value as
the approachability. The manipulability and approachability
are computed following these routines and are used to optimize the position
of handling. The rotation matrices, the manipulability, and the approachability
are shown in Fig.\ref{discretization}.
The left part of Fig.\ref{discretization} shows the rotation matrices.
The purple vectors are the approaching directions
and the rotation matrices are
sampled around the approaching directions.
The right part of Fig.\ref{discretization}
shows the manipulability and approachability.
The manipulability at the grid along different approaching directions is
rendered with a color spectrum in the the-lighter-the-higher style. The
approachability is rendered by the number of vectors at each grid.

We choose the grid position that has largest approachability as the optimized
position for handling. This is because at this position the two hands can
approach an object with a large number of directions as well as high
manipulability. The optimized position will be the translational component of
the handling configuration.

The rotational component of the handling configuration is evenly
sampled. We compare different heuristic strategies
for the rotational component and conclude that even sampling is
the most effective one. The second box in Fig.\ref{stateplan}(b)
shows the evenly sampled rotations. The left figure shows the object
at the first rotation matrix. The middle one shows both the object at
the first and second rotation matrices. The right figure shows
the overlap of all rotation.

Each position and rotation pair is a candidate handling configuration and we
associate it with the grasps of both hands. This is done on-line by checking the
collisions between the hands and their approaching directions (the hand
directions that induce crossing arms are also removed in this
step). Meanwhile, we rate the quality of each handling configuration using its
angular distance to the goal. The handling configuration that has a shorter
distance to the goal is set with higher priority and will be tried
first. More exactly, we loop the handling configurations using the
second level of discretization during the handling trial.
That is, the most outside loop is the approaching vectors.
We try the nearest handling configuration to the goal at each approaching vector
in the first loop, try the second-nearest handling configuration in
the second loop, and so on. The looping process will be further discussed
in Section V.B.

\section{Building and Searching the Regrasp Graphs}

\begin{figure}[!htbp]
  \centering
  \includegraphics[width = 3.2in]{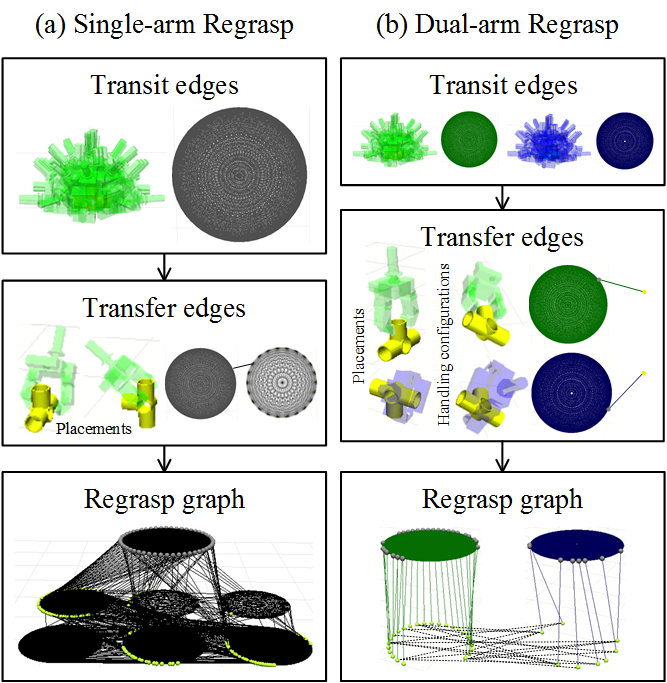}
  \caption{Building the regrasp graphs. In the single-arm case,
  we first arrange the grasps associated with the same placement around
  a circle, and make them fully connected (top box). 
  Then, we connect two circles by checking if they
  share a common grasp (e.g. the same grasp described in the
  object's local coordinate system). Using these two
  steps, we get a graph like the one shown in the bottom box.
  The dual-arm case involves both placements and handling
  configurations. The placements are similar to the single-arm
  case and encode transit motion using fully connected circles
  (see the green and blue circles in top and middle boxes).
  The grasps associated with handling configurations are also arranged around
  circles, but they are not connected inside the circle.
  The lower layer in the bottom box shows this kind of circles.}
  \label{graph}
\end{figure}

Fig.\ref{graph} shows the flow charts of building the
regrasp graphs. The grasps associated with
the same state are connected using transit edges and
the grasps associated with different states are connected
using transfer edges. The difference between single-arm
regrasp graph and dual-arm regrasp graph is: In single-arm
case, the regrasp graph is built once and dynamically trimmed
using the results of searching. In dual-arm case, the regrasp
graph is built dynamically using the selected handling configuration.

\subsection{Single-arm case}

In single-arm case, we only build the regrasp graph for one time.
First, we arrange the grasps associated with the same state
around a circle and make them fully connected. The edges
inside the circle are transit type since the grasp at one
end of the edge can be changed to the grasp at the other end
without moving the object. The figures in the top box of Fig.\ref{graph}(a)
illustrates one of the arrangement and connection. The circle on
the right corresponds to the grasps and placement on the left.
Then, we connect the circles. If two grasps associated with different
states are the same in the object?s local coordinate system, we
add an edge between them. The edges connecting the circles are
transfer type since the grasp at one end of the edge can be changed
to the grasp at the other end by moving the object. Finally,
we get a regrasp graph like the figure in the bottom box. Each
circle represents one state, and each node on the circle represents
one grasp. Edges inside the circle represents transit motion
and edges connecting the circles represent transfer motion.
The grasp has two layers where the upper one only includes the initial
placement and the lower layer has all possible placements. This two-layer
structure enables both simple pick-and-place planning, namely pick-and-place
with the same placement, and reorientating planning that picks and places at
different placements.

\begin{figure}[!htbp]
  \centering
  \includegraphics[width = 3.2in]{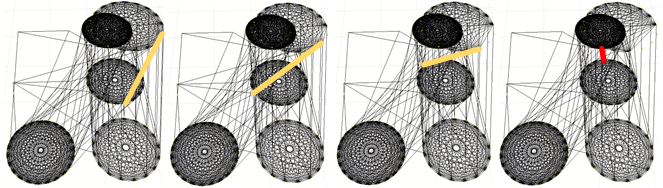}
  \caption{Searching the single-arm regrasp graph.
  The single-arm regrasp is built once but the edges and nodes
  are dynamically trimmed during searching and motion planning.
  The first three paths (rendered in yellow)
  include unavailable nodes or edges.
  They are deleted and searched again.}
  \label{sglsearch}
\end{figure}

Given the initial placement and goal placement, we search
the regrasp graph to find a sequence of pick-and-place operations.
The starting node of searching is a random grasp on the circle
in the upper layer and the ending node
is a random grasp on the circle that corresponds to the end
placement in the lower layer.
The result of searching is a sequence of states,
grasps, and transit and transfer motion connecting the
starting and ending nodes. For each state and grasp in the
searched result, we check if they are IK-feasible and
collision-free. In the placement planning part, we only
check the collisions between the hand, the object, and
the table. The IK solving and the collision
detection between the robot and other obstacles in the
environment are delayed to this searching step. If the sequence is
IK-infeasible or collides with obstacles, we trim the regrasp
graph by removing the correspondent grasps and search again.
Or else, we do motion planning using Transition-based RRT\cite{Jaillet08}
for the edges in the searched result. Like the grasps, we
trim the regrasp graph by removing the correspondent edges
and search again if motion planning fails. Fig.\ref{sglsearch} shows this
process. Either an available sequence is found or the regrasp
graph becomes unconnected after the search.

\subsection{Dual-arm case}

In dual-arm case, we loop the rated handling configurations and
rebuild the regrasp graph at each loop. The regrasp graph
is composed of two sub-graphs. One for the master hand and
one for slave hand. Each sub-graph is composed of only two states:
One is the initial or goal placement, the other is the handling
configuration. The green and blue plots in the bottom box of Fig.\ref{graph}(b)
show the two sub-graphs. The upper layer of the green plot is
the initial placement and the upper layer of the blue plot is
the goal placement. They are built in the same way as the
placement circles in the single-arm case.
Instead of delaying IK-solving and
collision detection to the searching step, we do them before
starting the loop:
Only the available grasps are used to build regrasp graphs. 
The lower layers of the plots are the
handling configurations. The two sub-graphs share a same handling
configuration and are bridged at it.
The edges connecting
the upper layer and the lower layer indicate the transfer motion.
If a grasp associated with the handling configuration and a grasp
associated with the initial or goal placement are the same in
the object?s coordinate system, we connect them with a transfer
edge.

\begin{figure}[!htbp]
  \centering
  \includegraphics[width = 3.2in]{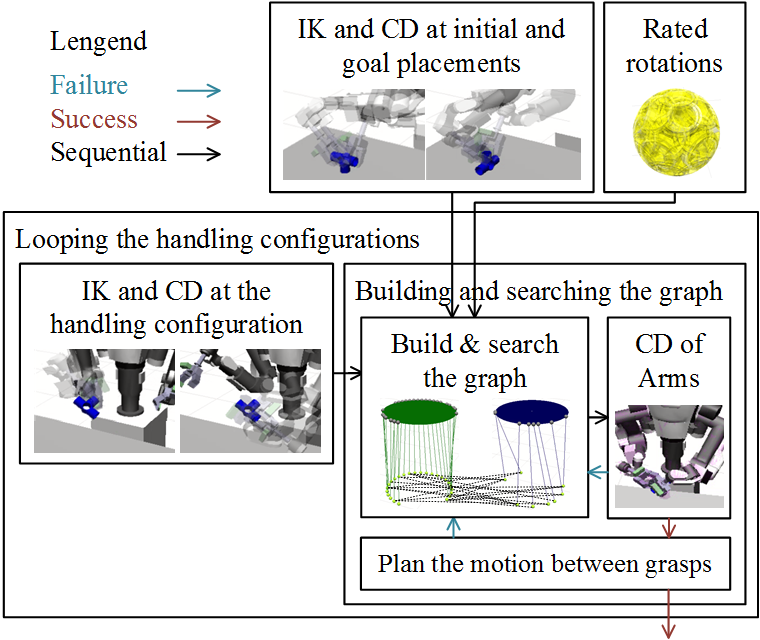}
  \caption{Searching the dual-arm regrasp graph.
  In the ``looping the handling configurations'' box,
  we loop the handling
  configurations and build the regrasp graph.
  The graph is rebuilt for each handling configuration.
  In the ``building and searching the graph'' box,
  we use the same searching method shown in Fig.\ref{sglsearch}.
  Checking the IK and collision at initial and goal placements and rating the
  the rotation of handling configurations are
  performed before starting the loop.
  }
  \label{dualsearch}
\end{figure}

At the handling configuration, the robot cannot perform transit
motion with a single arm and there is no edges inside the
circle. But the robot can transit bi-manually: It
could hold the object with one hand and perform transit motion
with the other hand. Therefore, we connect the two
circles at the lower layers of the sub-graphs using transit
edges, which are illustrated with dotted lines in
Fig.\ref{graph}. The nodes on the lower circles and the edges
connecting them change as we select different configurations and
rebuild the regrasp graphs.
IK-solving and collision detection at the nodes on the lower
circles are performed each time the regrasp graph is rebuilt.
During searching, we no long need to waste time on IK. However,
further checking the collision between master and slave arms is required.
The collision detection done during rebuilding the graph was performed
independently on two arms. The collision between master and slave arms
need to be further checked before merging the two sub-graphs.
Fig.\ref{dualsearch} shows the process.

\section{Experiments and Analysis}

We analyze the performance of our algorithms using simulation.
The model of our objects are based on an ``L''-shape block,
a ``box"-shape part, and a ``T''-shape tube (see
Fig.\ref{statistical}).
Our robot platform is the HiroNX\footnote{See http://nextage.kawada.jp/en/} and
our simulation software is based on Choreonoid and
its graspPlugin\footnote{See http://choreonoid.org/en/,
http://choreonoid.org/GraspPlugin/i/?q=en}.
Real-world executions of single-arm regrasp using 
KINECT were published in our previous work\footnote{See a video at
http://youtu.be/Mlgl8BmvfPc}.
This paper concentrates on simulation and the comparison of single-arm
and dual-arm regrasp. It doesn't present the
details of those implementations.

\subsection{Respective analysis}

\begin{figure*}[!htbp]
  \centering
  \subfigure[The snapshots of single-arm regrasp. In (1), the robot does
  motion planning to grasp the object. The virtual robot sequences drawn in
  purple are the key frames of the planned motion. It belongs to the transit
  motion type. On the regrasp graph (see the second row), it is a single node
  in one circle (marked with red color). In (2), the robot finds the initial and
  goal of a transfer motion. The virtual robot drawn in yellow color shows the
  initial and the virtual robot drawn in cyan color shows the goal.
  The robot does motion planning to
  perform this transfer motion in (3). On the regrasp graph, (2) and (3)
  correspond to an edge connecting two circles. In (4) and (5),
  the robot ``regrasps'' the object by changing
  the grasps. The motion in these figures is transit. They correspond to an edge
  in the same circle on the regrasp graph.
  (6) and (7) is the same as (2) and (3), and they are the task and
  motion planning result of the second transfer motion. They correspond to
  an edge connecting two circles. In (8), the robot does the final motion
  planning to retract the hand to standard pose. It is a single node on the
  regrasp graph.]{
    \includegraphics[width=6.7in]{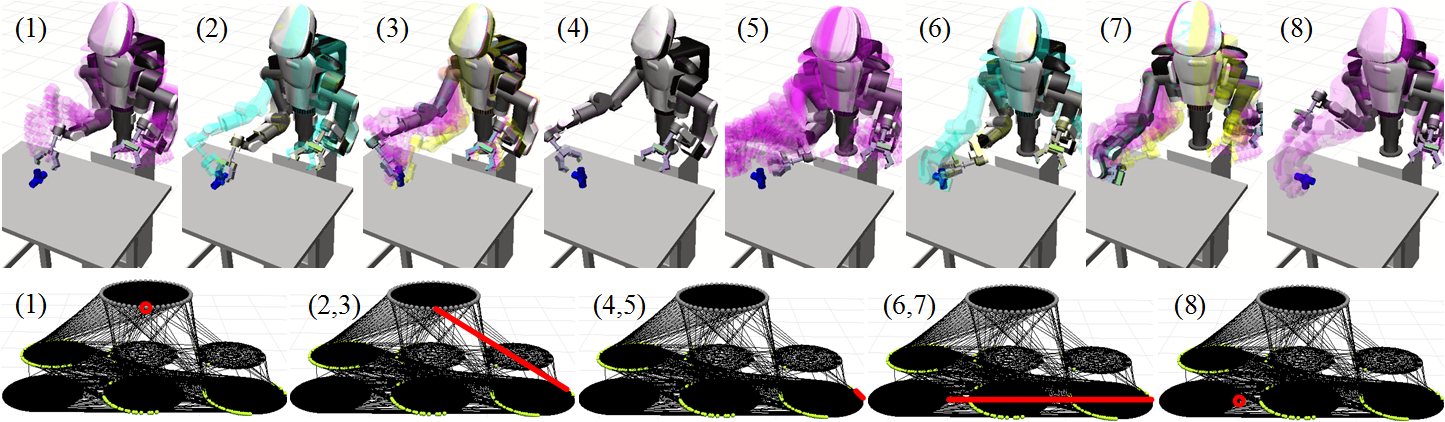}
    \label{fig:snapshot_a}}
  ~
  \subfigure[The snapshots of dual-arm regrasp. In (1), the robot does
  transit motion planning to grasp the object with master hand. It corresponds
  to a node in the upper circle (see the lower row of figures). In (2)
  and (3), the robot transfers the object into a handling configuration with
  master hand. It corresponds to an edge connecting
  the upper and lower layers in the master sub-graph. In (4), the robot
  ``regrasp'' the object with the slave hand. The motion is transit and
  corresponds to an edge connecting the lower circles of the two sub-graphs.
  In (5), the robot retract the master hand. In (6) and (7), the robot
  transfers the object into the goal placement on the table with the slave hand.
  It corresponds to an edge connecting the lower and upper layers in the slave
  sub-graph. Finally, the robot transit its slave hand back in (8). Like (a),
  when the robot does motion planning, we draw a sequence of virtual robots in
  purple color. When the robot does transfer motion, we draw the initial and
  goal virtual robots in yellow and cyan colors.]{
    \includegraphics[width=6.7in]{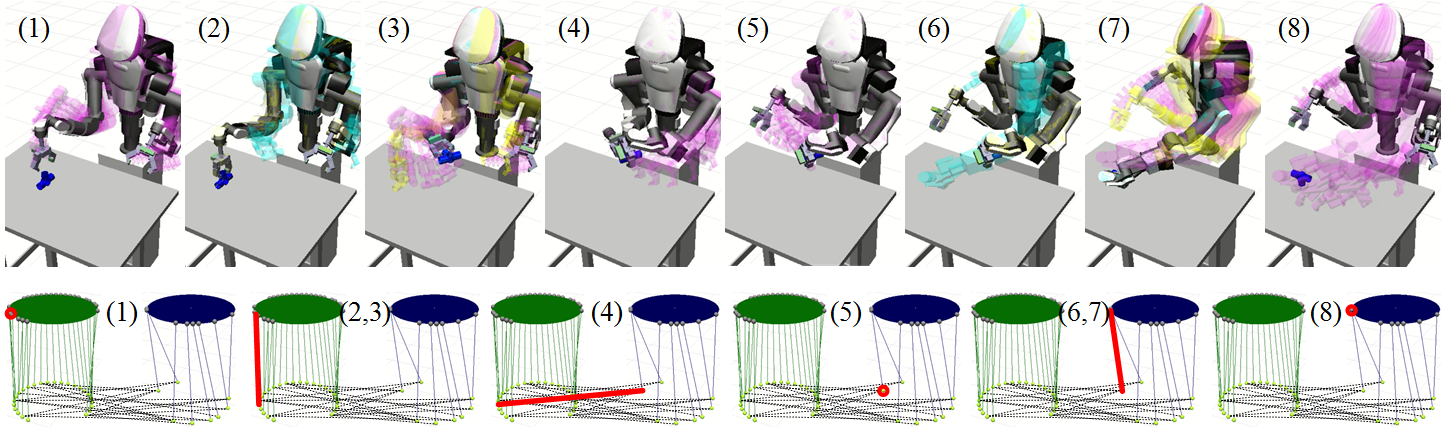}
    \label{fig:snapshot_b}}
  ~
  \caption{Snapshots of the robot regrasping the ``T''-shape tube using
  single-arm and dual-arm regrasp in simulation.}
  \label{snapshots}
\end{figure*}

Fig.\ref{snapshots} shows the snapshots of single-arm and dual-arm
regrasp. The initial configuration of these results is an object placing on a
table, and the goal is the object placing at the same place with a different
placement. The robot is required to reorientate the object with either
single-arm or dual-arm regrasp. We draw the key frames of the transit and
transfer motions in the upper row of each sub-figure, and draw the
correspondent regrasp graph in the lower row. The correspondence between the
nodes, edges, and the key frame figures are marked with red circles and
segments, and are written on the upper-left corner of each sub-figure in the
lower row (See the caption for details).
The results of other objects are shown in the video attachment.

\begin{figure}[!htbp]
  \centering
  \includegraphics[width = 3.4in]{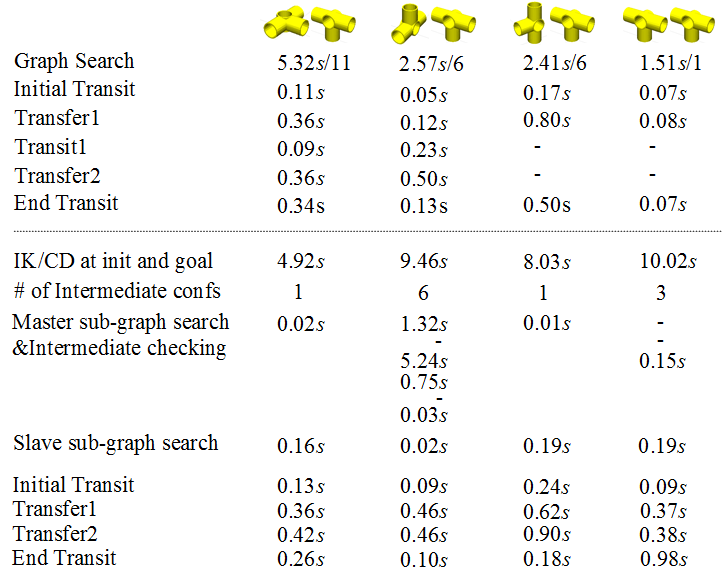}
  \caption{Time cost of graph search and motion planning in
single-arm (above the dashline) and dual-arm regrasp (below the dash line).}
  \label{cost}
\end{figure}

Fig.\ref{cost} shows the time cost of graph search and motion planning
in single-arm and dual-arm regrasp. Each column of
the table is the result of one regrasp task where the initial
and goal placements are drawn in the top. The figure is divided into two parts
by a dash line where the upper part is single-arm regrasp. The time cost of
singl-arm graph search and its number of research times are shown in the first
data row. The time cost of motion planning are shown in the other data rows.
Graph search takes less than 10$s$ and an average of 6 times of re-search for
this object. Motion planning is less than 1$s$ for both transit and transfer
motion. The results of simple pick-and-place task where no reorientation is
required are shown in the last column. It takes 2$s$ for graph search and less
than 0.1$s$ for motion planning.
The lower part of Fig.\ref{cost} is the cost of dual-arm
regrasp. The first row shows the time cost of IK solving and collision detection
at initial and goal placements. The second row shows the number of rotation tried
during searching. The first two rows correspond to the top two boxes in
Fig.\ref{dualsearch}. The third row shows the time cost of searching the master
sub-graph, and the fourth row shows the time cost of searching the slave
sub-graph. Depending on the number of intermediate configurations tried, the
number of data in the third row changes. The rows 5-8 are the time cost of
motion planning. The most time-consuming part is the IK solving and collision
detection at initial and goal placements. It is 10.02$s$ in the worst case.
The time cost is well acceptable consdiering that we sample the object surface
using 0.01$m$ granularity and sample the approaching directions at every
$\frac{\pi}{4}$$radian$. The second time-consuming part is searching the master
graph. Since we build the graph for each handling configuration, we might
test all possible rotations which costs as much as 6 $minutes$
in the worst case. This estimation is based on 352 rotations and 1$s$ time
cost per search. During the experiments, none of the search
exceeds 3$minutes$. The algorithms are fast enough to deal with large number
of grasps and exploded combinatorics and can be run thousands of times for
comparison.

\subsection{Comparison}

In order to compare single-arm and dual-arm regrasp, we set the initial and goal
of the object at a fixed position and use a random rotation around the normal of
the table surface to generate initial and goal placements. The robot is required
to reorientate the object from the initial placement to the goal using
single-arm or dual-arm regrasp.

\begin{figure*}[!htbp]
  \centering
  \includegraphics[width = 6.95in]{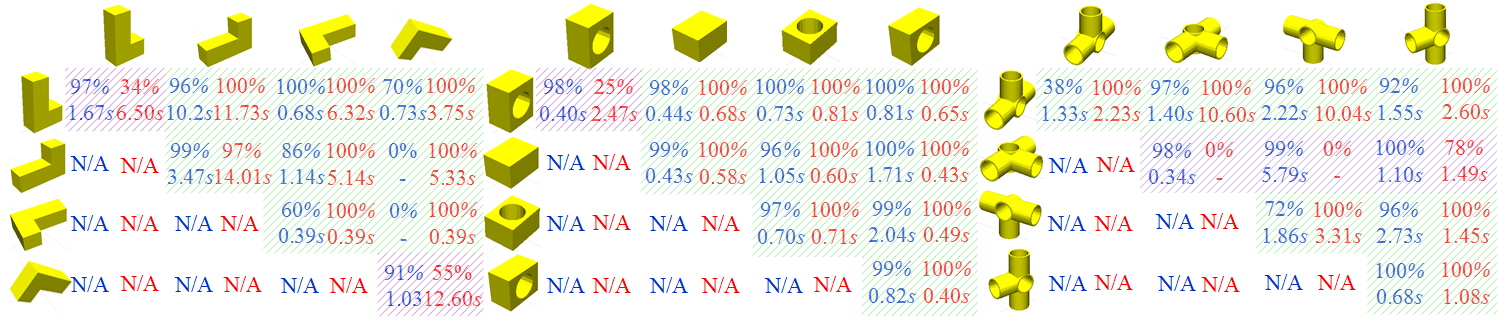}
  \caption{Average successful rates and costs of single-arm and dual-arm
  regrasp. Dual-arm regrasp has higher performance at the tasks under the
  green shade, and has lower perfomrance at the tasks under the purple shade.}
  \label{statistical}
\end{figure*}

Fig.\ref{statistical} shows the results of the comparison. The left column of
each table lists the initial placements and the upper row lists the goal
placements. The positions are fixed but we set a random yaw angle to the
initial placement to increase uncertainty. At each grid, the blue data shows
the successful rate and time cost of single-arm regrasp. In constrast, the red
data shows those of the dual-arm regrasp.
Like our expectation, dual-arm regrasp has good performance in most cases (see
the area shaded in green color in Fig.\ref{statistical}). It is sometimes a bit
slow, but the extra time cost is acceptable and the successful rate is much
higher than single-arm regrasp.
In a few cases, however, dual-arm regrasp has bad
performance (see the area shaded in purple color).
This usually happens to the simple pick-and-place tasks where no reorientation
is needed: See (row 1, column 1) and (row 4, column 4) of the ``L''-shape tube,
(row 1, column 1) of the ``box''-shape part, and (row 2, column 2) of the
``T''-shape tube. In these cases, the grasps of the master hand and the slave
hand are on the upper part of the object to ensure it can be placed down to the
same placement. They overlap with each other and collide
during handling.
(row 2, column 3) and (row 2, column 4) of the tube are also difficult to
dual-arm regrasp. The reason is similar -- the grasps of the master hand at the
initial placement overlap with the grasps of the slave hand at the goal
placement, and the master hand and the slave hand collide during handling. The
results demonstrate that dual-arm regrasp is 
not necessarily better than single-arm regrasp. The performance depends on
object shapes and the overlapping of grasps.

\section{Conclusions and Future Work}

In this paper, we develop efficient algorithms for single-arm and dual-arm
regrasp and run the algorithms thousands of times to compare their performance.
We confirm the algorithms are fast to deal with large number of grasps
and find dual-arm regrasp is not necessarily better than single-arm regrasp:
It depends on object shapes and the overlapping of grasps. We expect the results
will help practitioners to choose proper number of arms.

In the future, we would
further explore the features to choose the arm numbers using deep learning and
develop algorithms to predict whether a robot should use a single arm or dual arms.

\bibliographystyle{ieeetr}
\bibliography{references}

\end{document}